\def\paperID{13763}
\def\confName{CVPR}
\def\confYear{2024}

\def\paperTitle{Amodal Completion via Progressive Mixed Context Diffusion}

\def\authorBlock{
    Katherine Xu$^1$ \qquad
    Lingzhi Zhang$^2$ \qquad
    Jianbo Shi$^1$ \\
    $^1$University of Pennsylvania, $^2$Adobe \\ \\
}

\newif\ifreview 
\newif\ifarxiv \newcommand{\arxiv}{\arxivtrue}
\newif\ifcamera 
\newif\ifrebuttal

\arxiv

\pdfoutput=1
\documentclass[10pt,twocolumn,letterpaper]{article}
\ifreview \usepackage[review]{cvpr} \fi
\ifarxiv \usepackage[pagenumbers]{cvpr} \fi
\ifrebuttal \usepackage[rebuttal]{cvpr} \fi
\ifcamera \usepackage{cvpr} \fi

\usepackage{graphicx}	
\usepackage{amsmath}	
\usepackage{amssymb}	
\usepackage{booktabs}
\usepackage{times}
\usepackage{microtype}
\usepackage{epsfig}
\usepackage[table,xcdraw,dvipsnames]{xcolor}
\usepackage{caption}
\usepackage{float}
\usepackage{placeins}
\usepackage{color, colortbl}
\usepackage{stfloats}
\usepackage{enumitem}
\usepackage{tabularx}
\usepackage{xstring}
\usepackage{multirow}
\usepackage{xspace}
\usepackage{url}
\usepackage{subcaption}
\usepackage{xcolor}
\usepackage[hang,flushmargin]{footmisc}

\usepackage{algorithm}
\usepackage{algpseudocode}
\algnewcommand{\LeftComment}[1]{\Statex \(\triangleright\) #1}
\usepackage{wrapfig}
\usepackage{tabularray}

\ifarxiv  \fi

\definecolor{redpath}{rgb}{0.752, 0, 0}
\definecolor{bluepath}{rgb}{0, 0.125, 0.773}
\definecolor{purplepath}{rgb}{0.596, 0, 0.710}
\definecolor{greenpath}{rgb}{0.059, 0.710, 0.216}

\usepackage{xr-hyper}

\makeatletter
\newcommand*{\addFileDependency}[1]{
  \typeout{(#1)}
  \@addtofilelist{#1}
  \IfFileExists{#1}{}{\typeout{No file #1.}}
}

\makeatother

\definecolor{cvprblue}{rgb}{0.21,0.49,0.74}
\usepackage[pagebackref,breaklinks,colorlinks,citecolor=cvprblue]{hyperref}
\usepackage[capitalize]{cleveref}
\crefname{section}{Sec.}{Secs.}
\crefname{table}{Table}{Tables}
\crefname{figure}{Fig.}{Figs.}

\begin{document}
\title{\paperTitle}
\author{\authorBlock}

\twocolumn[{%
\maketitle
\vspace{-32 pt}
\begin{center}
    \centering
    \includegraphics[trim=0in 3.45in 2.1in 0in, clip,width=\textwidth]{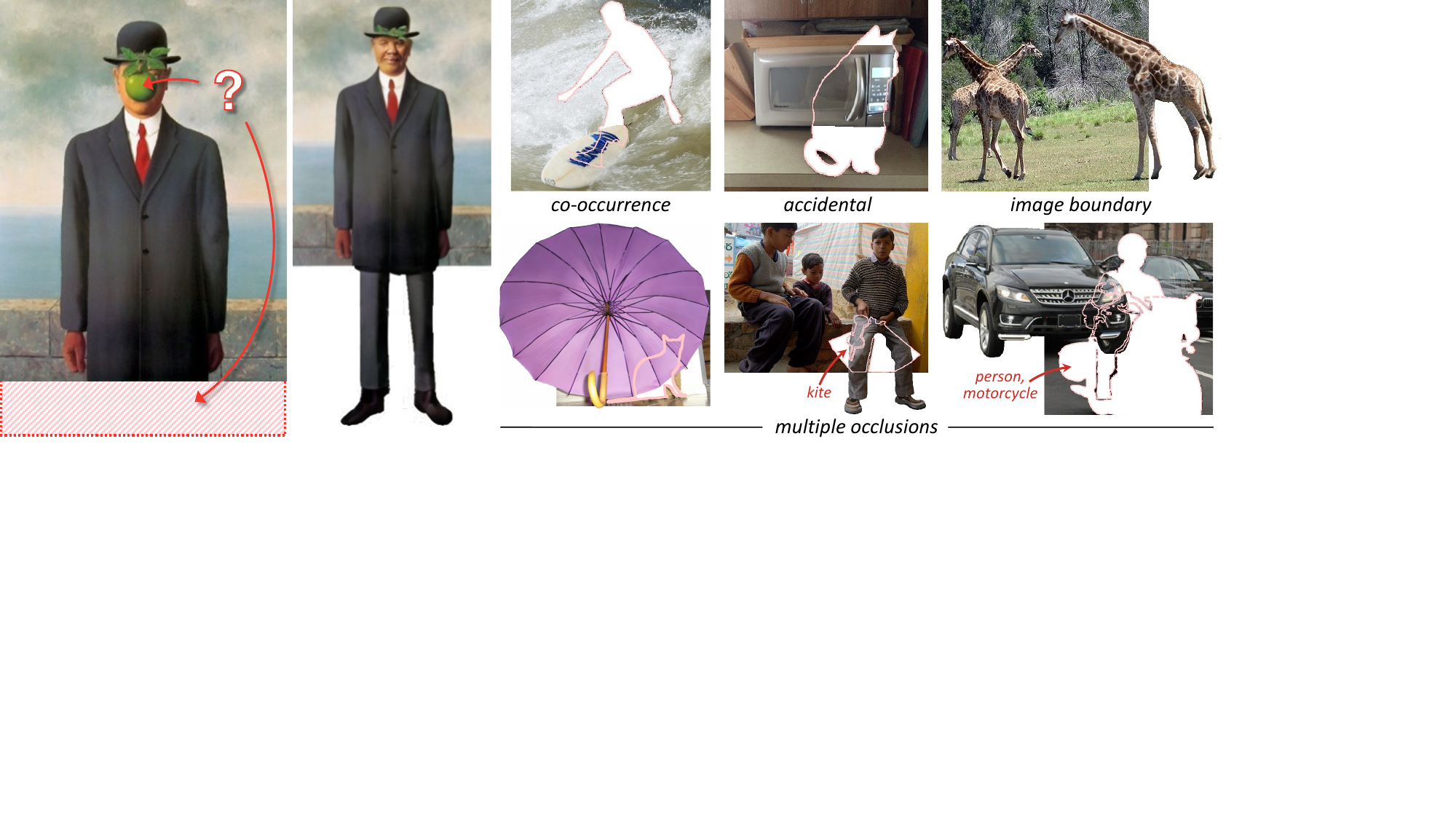}
    \vspace{-20 pt}
    \captionof{figure}[Teaser]{Our method can recover the hidden pixels of objects in diverse images. Occluders may be co-occurring (a person on a surfboard), accidental (a cat in front of a microwave), the image boundary (giraffe), or a combination of these scenarios.}
    \label{fig:teaser}
\end{center}%
}]

\begin{abstract}
\vspace{-11 pt}

Our brain can effortlessly recognize objects even when partially hidden from view. Seeing the visible of the hidden is called amodal completion; however, this task remains a challenge for generative AI despite rapid progress. We propose to sidestep many of the difficulties of existing approaches, which typically involve a two-step process of predicting amodal masks and then generating pixels. Our method involves thinking outside the box, literally!  We go outside the object bounding box to use its context to guide a pre-trained diffusion inpainting model, and then progressively grow the occluded object and trim the extra background. We overcome two technical challenges: 1) how to be free of unwanted co-occurrence bias, which tends to regenerate similar occluders, and 2) how to judge if an amodal completion has succeeded. Our amodal completion method exhibits improved photorealistic completion results compared to existing approaches in numerous successful completion cases. And the best part? It doesn't require any special training or fine-tuning of models.

\end{abstract}
\vspace{-18pt}

\section{Introduction}
\label{sec:intro}
\vspace{-6pt}

Have you ever wondered how objects regularly occlude one another, yet we can effortlessly recognize and imagine their unoccluded appearance? Our visual system performs this task of \textit{amodal completion} using the continuity and symmetry of an object's shape \cite{Lier1999InvestigatingGE} and everyday familiarity of the world \cite{yun2018temporal}. Amodal completion, filling in hidden object parts, is a challenging AI task despite rapid advances in computer vision. This technology has many applications, such as in robotics, autonomous vehicles, and augmented reality.

A reasonable amodal completion approach \cite{zhan2020selfsupervised, Bowen2021OCONetIE} consists of two stages: 1) completing a binary amodal mask; 2) synthesizing RGB pixel values within the mask. However, directly regressing the amodal mask is an ill-posed formulation due to the diversity of possible completions.

Computational issues aside, how can we create a dataset for amodal completion?  Previous research attempts to construct datasets through random mask placement to simulate occlusion \cite{li2016amodal}, computer graphics rendering techniques \cite{hu2019sail}, or by asking humans to label amodal segmentation masks given the modal masks \cite{zhu2017semantic, kins, ao2023amodal}. However, a domain gap persists between synthetic and natural images, and labeled natural images are expensive to obtain.

But what if we can sidestep all these difficulties?  This involves thinking outside the box, literally: 1) extend from the bounding box of an occluded object to include sufficient image context, 2) remove the occluders, 3) use a pre-trained diffusion model to grow the object, and 4) trim off the extra background. Our amodal completion pipeline avoids predicting the amodal segmentation mask as an intermediate step. Furthermore, we can recover occluded pixels within occluder objects and \emph{beyond the image boundary} (Figure \ref{fig:teaser}), generate \emph{diverse} versions of completed pixels, and require \emph{no training or fine-tuning} of the diffusion model.

Straightforward usage of an off-the-shelf diffusion model for image inpainting succeeds only sometimes. Failure cases often generate other objects within the occluder masks: original occluder look-alikes or unintended things that co-occur with the object of interest. Imagine removing a hand holding a cup; diffusion inpainting often adds a different hand simply because we don't see a floating cup in real life.

\emph{How can we add control to discourage the pre-trained diffusion model from re-generating co-occurrence?}   Two tasks exist to solve: momentarily breaking free of context, and knowing when amodal completion has succeeded.

First, to break free of the contextual bias that causes co-occurrence, 
we propose \emph{mixed context diffusion sampling} to temporarily replace the image context with a natural clean background, akin to product photography. We intercept the diffusion process halfway, extracting a pseudo-complete object in a still-noisy image using unsupervised clustering of decoder features.  Then, we use the pseudo-complete object as a reference target to the reverse diffusion process while gradually reintroducing the original image background.

Second, to infer whether the amodal completion succeeds, we introduce counterfactual reasoning: use the generated object and outpaint its background.  If the object is complete, then any outpainting should not increase its size.  Thus, we can judge whether amodal completion is successful by comparing the object segmentation before and after outpainting.

With these two tools, we progressively run a pre-trained text-based diffusion inpainting model \cite{rombach2022ldm} until the occluded object is complete. Our method requires no extra datasets, re-training, or adaptation. It is entirely based on pre-trained diffusion models \cite{rombach2022ldm}, complemented by off-the-shelf grounded segmentation \cite{liu2023groundingdino, kirillov2023segment_sam} and depth models \cite{lee2022instance} as auxiliary modules. In summary:

\begin{enumerate}[noitemsep, nolistsep]
    \item We introduce a \emph{progressive occlusion-aware amodal completion pipeline} that effectively recovers hidden pixels within occluder masks and beyond the image boundary.
    \item As a pioneering exploration, we identify the challenge of a diffusion inpainting model generating unwanted co-occurring objects, instead of completing objects. We propose \emph{mixed context diffusion sampling}, which modifies the image context to overcome difficult co-occurrence.
    \item We create a training-free \emph{counterfactual completion curation system} to decide if a generated object is complete.
\end{enumerate}

\vspace{-6pt}
\section{Related Work}
\label{sec:related}
\vspace{-6pt}

\textbf{Amodal completion.} Amodal appearance completion aims to fill in the hidden regions of occluded objects.  Current methods often rely on a two-step approach of first predicting the amodal mask and then generating the object appearance given the amodal mask.  These approaches have been applied on toy datasets \cite{burgess2019monet, greff2020multiobject, engelcke2020genesis} and specific object categories such as vehicles \cite{yan2019visualizing, variational_amodal, zheng2021visiting_csdnet}, humans \cite{zhou2021humandeocc}, and food \cite{papadopoulos2019make_pizzagan}. Additional methods perform amodal completion for common object categories, mainly on synthetic indoor scenes \cite{ehsani2018segan, dhamo2019objectdriven, zheng2021visiting_csdnet}. However, there is a domain gap between synthetic and natural images, and so we are interested in generating the amodal completion of common objects in natural scenarios. Zhan et al. \cite{zhan2020selfsupervised} and Bowen et al. \cite{Bowen2021OCONetIE} perform amodal completion in natural images, but these GAN-based methods tend to lack high image fidelity. In contrast, our approach leverages the good image prior of pre-trained diffusion models to photorealistically complete objects in natural images.

An alternative means of tackling the amodal completion task is directly inpainting occluder regions, such as by using large mask image inpainting \cite{suvorov2021lama} or training diffusion inpainting models to remove objects \cite{yildirim2023instinpaint}. There is also a line of research in image outpainting \cite{cheng2022inout, xiao2020image_out, Li_2022_CVPR_context_out, wang2022structureguided_out, Wang_2019_CVPR_wide_context_out, Li_2021_WACV_controllable_progressive_out}. However, since these approaches are not meant for amodal completion, they often produce realistic images but fail to complete the appearance of desired objects under significant occlusion. In contrast, our method can realistically generate the amodal completion of occluded objects by progressively inpainting occluder regions and disentangling co-occurrence bias.

\textbf{Diffusion models.} Inspired by non-equilibrium thermodynamics \cite{sohldickstein2015deep}, diffusion models achieve remarkable results for text-to-image and image inpainting tasks \cite{nichol2022glide, ramesh2022hierarchical, rombach2022ldm, saharia2022imagen}, often outperforming GANs in generating photorealistic and diverse images \cite{dhariwal2021diffusion, ho2020denoising, nichol2021improved}. However, these approaches provide limited control of image generation outside the text prompt. Several works provide additional guidance to diffusion models for controllable image generation using CLIP \cite{nichol2022glide, ramesh2022hierarchical}, cross-attention and self-attention \cite{hertz2022prompt_to_prompt, patashnik2023localizing, chen2023trainingfree, feng2023trainingfree}, stroke paintings \cite{meng2022sdedit}, exemplar images \cite{yang2022paint}, and extra conditioning on segmentation maps, edge maps, bounding boxes, and keypoints in ControlNet \cite{zhang2023adding}. However, employing these techniques for amodal completion typically needs an amodal mask or `complete' edge map as guidance, which is not available. Moreover, they often require resource-intensive re-training of the diffusion models. In contrast, our method does not assume any initial guidance and is entirely training-free.

\vspace{-1pt}

\vspace{-6pt}
\section{Method}
\label{sec:method}

\subsection{Preliminaries}
\label{sec:method_background_diffusion}

Diffusion models \cite{ho2020denoising} learn a data distribution $p(I)$ using a sequence of denoising. In the forward process, the model adds noise to an image \(I\) in $N$ time steps, resulting in the sample having approximately Gaussian noise. In the reverse process, the model learns to denoise the sample in $N$ steps. At each step $t = [1, N]$, a learned neural network predicts the noise $\epsilon_\theta(I^t, t)$ given the noisy image $I^t$.

Unlike diffusion models that use the image pixel space, latent diffusion models (LDMs) \cite{rombach2022ldm} operate in the latent space of pre-trained autoencoders.
For an image $I \in \mathbb{R}^{H \times W \times 3}$, the encoder $E$ encodes \(I\) into the latent representation $\mathbf{z} = E(I)$, and the decoder $D$ reconstructs \(I\) from $\mathbf{z}$ using $\hat{I} = D(\mathbf{z})$.
In the diffusion process, the autoencoder can be viewed as a time-conditional UNet \cite{ronneberger2015unet}, $\epsilon_{\theta}(\mathbf{z}_t, t)$, for a given time $t$ and latent $\mathbf{z}_t$. To incorporate any input condition $y$ such as an inpainting mask, LDMs add cross-attention layers \cite{vaswani2017attention} to the denoising UNet so that $y$ maps to the intermediate layers of the UNet \cite{rombach2022ldm}.
In this work, we use the publicly released Stable Diffusion v2 inpainting model checkpoint \cite{rombach2022ldm}. For simplicity, \emph{our notation uses the image pixel space hereafter}.

\begin{figure*}[!ht]
    \vspace{-15 pt}
    \centering
    \includegraphics[trim=0in 4.75in 0.3in 0in, clip,width=\textwidth]{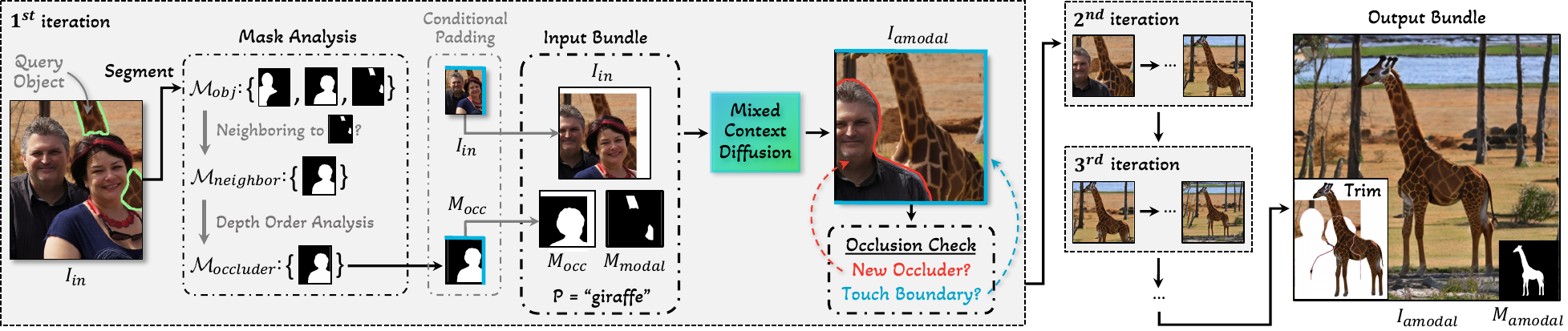}
    \vspace{-18 pt}
    \caption{Our \textbf{Progressive Occlusion-aware Completion} pipeline. \textbf{First iteration:} We perform instance segmentation \cite{liu2023groundingdino, kirillov2023segment_sam} and analyze the object masks to determine occluders \cite{lee2022instance}. If the query object touches the image boundary, then we pad the image and mask to enable object completion beyond the boundary in those directions.  Using this input bundle, we run our Mixed Context Diffusion Sampling to obtain a new amodal completion image.  \emph{The details of Mixed Context Diffusion Sampling are in Figure \ref{fig:mixed_context_sampling}.}   Next, we check whether the generated object has a new occluder or touches the image boundary. In this example, the man from the original image appears as a new occluder that was previously undetected. \textbf{Additional iterations:} If the query object remains occluded, then we run additional iterations of our pipeline. \textbf{Output:} We return the final amodal completion image and amodal mask, and we can trim extra background to overlay on the original image.}
    \label{fig:method_diagram}
    \vspace{-12 pt}
\end{figure*}

\subsection{Problem Setup}

The task of amodal completion entails identifying both visible and hidden aspects of objects, inside and outside an image's boundary. Given an arbitrary image \(I_{in}\) in \(\mathbb{R}^{H \times W \times 3}\) and an object of interest (`query object') with its modal mask \(M_{modal}\) in \(\mathbb{R}^{H \times W}\), our objective is to predict the amodal completion image \(I_{amodal}\) in \(\mathbb{R}^{H' \times W' \times 3}\) and the corresponding amodal mask \(M_{amodal}\) in \(\mathbb{R}^{H' \times W'}\). We use $H'$ and $W'$ to indicate that the final amodal completion image may differ in size from the original image due to potential extensions beyond the image boundary. Furthermore, we denote the diffusion inpainting/outpainting process as \(F_{s \rightarrow e}\), where \(s\) is the starting timestep and \(e\) is the ending timestep. The text prompt is represented as \(P\), which is the semantic category of the query object for our proposed solutions. \(M_{in}\) and \(I_{out}\) signify the generic input mask and the generated output, respectively. This diffusion process can be expressed as:

\vspace{-8 pt}
\begin{equation}
I_{out} = F_{s \rightarrow e}(I_{in}, M_{in}, P)
\label{equ:problem_setup_diffusion_process}
\end{equation}

For amodal completion to be successful, it must fulfill three criteria. 1) The process should exclusively remove occluders without altering the image background, thereby avoiding overextension of the object. 2) It must ensure a complete representation of all object parts to avoid incompletion. 3) The completion must be contextually consistent, avoiding any physically implausible object configurations.  We evaluate the first two criteria by developing a dataset of unoccluded objects from natural images, and then artificially generating the pseudo-occluded versions. Contextual consistency is harder to quantify, so we perform a user study.

\subsection{Naive Outpainting Approach}
\label{sec:method_motivation}

A simple approach to amodal completion may assume that all pixels outside the query object's modal mask are occlusions, which are then subject to outpainting. This `Naive Outpainting' approach can be mathematically expressed as:

\vspace{-8 pt}
\begin{equation}
I_{amodal} = F_{0 \rightarrow N}(I_{in}, 1 - M_{modal}, P)
\label{equ:naive_sd}
\end{equation}
where $N$ is the total number of timesteps for the diffusion process, which is set to 50 in the DDIM scheduler \cite{song2020denoising}. We treat all masks as binary, so $1 - M_{modal}$ signifies everything exterior to the query object.

\begin{figure}[!h]
    \vspace{-7 pt}
    \centering
    \includegraphics[trim=0in 5.65in 7.5in 0in, clip,width=\columnwidth]{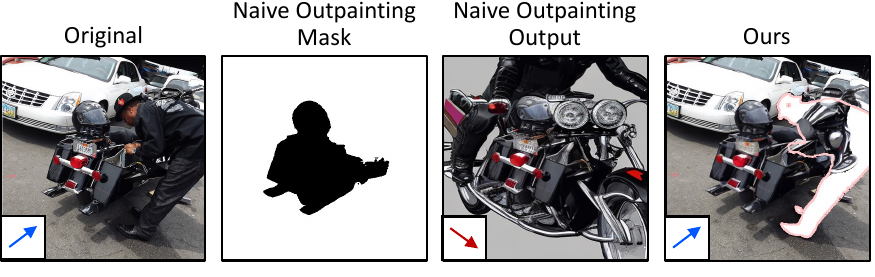}
    \vspace{-23 pt}
    \caption{Naively using a diffusion model to outpaint the query object may overextend the object and change its identity, such as the motorcycle changing orientation (indicated by arrow). In contrast, our method preserves the object's identity.}
    \label{fig:motivation_naivesd}
    \vspace{-13 pt}
\end{figure}

Naive Outpainting often overextends the query object due to the lack of contextual constraints, compromising the integrity of its identity and violating the objective of amodal completion. For example, this approach produces an unwanted change to the motorcycle's orientation in Figure \ref{fig:motivation_naivesd}.

\begin{figure*}[!ht]
    \vspace{-15 pt}
    \centering
    \includegraphics[trim=0in 2.4in 1.4in 0in, clip,width=\textwidth]{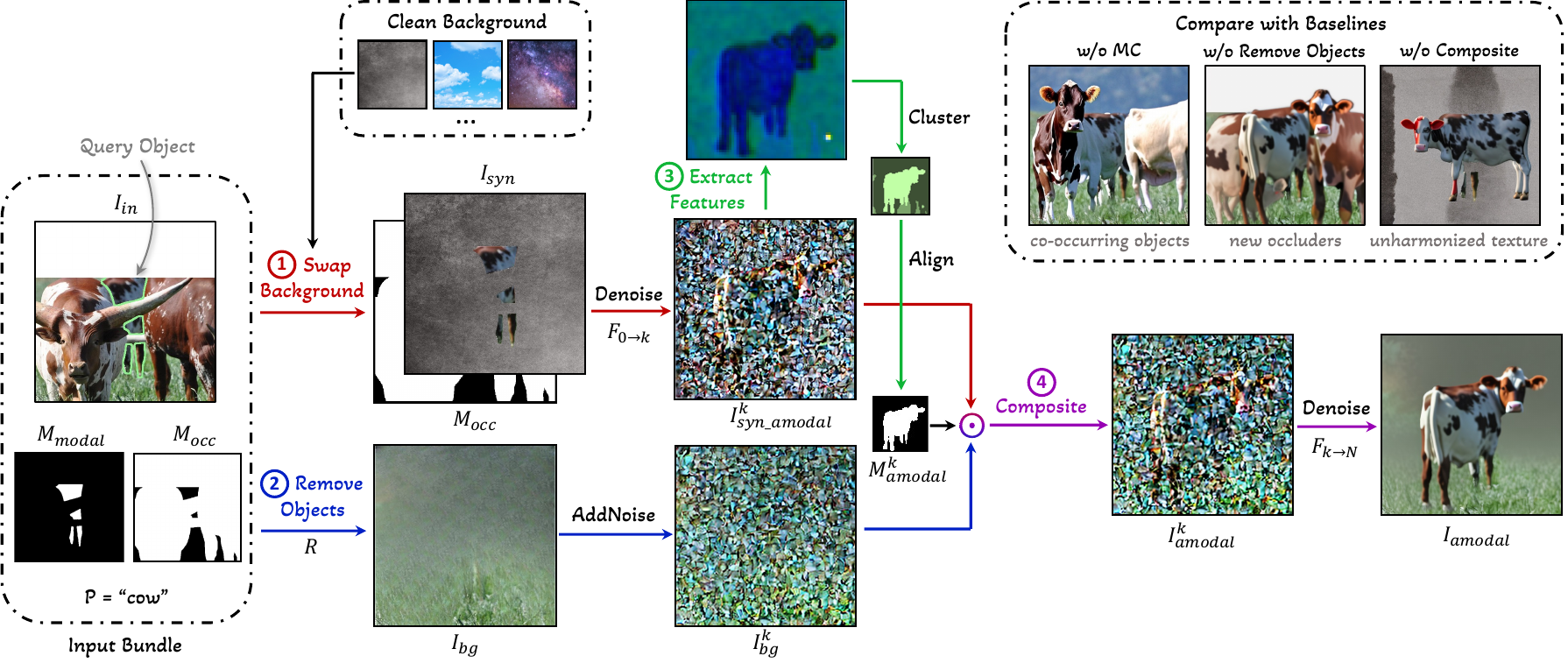}
    \vspace{-20 pt}
    \caption{Our \textbf{Mixed Context (MC) Diffusion Sampling}.     1) \textbf{\textcolor{redpath}{Swap background}} (\emph{red}): We replace the background of \( I_{in} \) using \( M_{occ} \) to create \( I_{syn} \), followed by diffusion inpainting to the \( k^{th} \) timestep, resulting in \( I_{syn\_amodal}^k \).     2) \textbf{\textcolor{bluepath}{Create object-removed background image}} (\emph{blue}): We remove query objects and occluders from \( I_{in} \) using a removal inpainter \cite{suvorov2021lama}, and then add noise up to the \( k^{th} \) timestep, producing \( I_{bg}^k \).     3) \textbf{\textcolor{greenpath}{Segment object in noisy image}} (\emph{green}): We extract diffusion features from \( I_{syn\_amodal}^k \), cluster them, and select the query object's amodal mask \( M_{amodal}^k \) at the \( k^{th} \) timestep by aligning with \( M_{modal} \). 4)  \textbf{\textcolor{purplepath}{Composite}} (\emph{purple}): We use \( M_{amodal}^k \) to place the query object from \( I_{syn\_amodal}^k \) onto the object-removed background image \( I_{bg}^k \). The final image, \( I_{amodal} \), is obtained by completing the remaining \( N-k \) diffusion steps, where \( N \) is the total number of steps. \textbf{Top right:} We show various failure cases if we remove parts of this MC method.}
    \vspace{-12 pt}
    \label{fig:mixed_context_sampling}
\end{figure*}

\subsection{Progressive Occlusion-aware Completion}
\label{sec:method_progressive_completion_pipeline}

Our method is based on two key insights: 1) inpainting only where necessary by identifying occluders prevents overextension, and 2) iteratively performing this inpainting step avoids incompletion. Thus, we propose a `Progressive Occlusion-aware Completion' pipeline, as shown in Figure \ref{fig:method_diagram}. Each iteration of our pipeline has several steps that are described below, and we perform more iterations if occluders remain. This approach significantly reduces object overextension, as evidenced by the results on the right side of Figure \ref{fig:experiments}.

\textbf{Mask analysis.}
Given the input image \( I_{in} \), the first step in each iteration is to identify all object masks by applying a grounded segmentation model \cite{liu2023groundingdino, kirillov2023segment_sam}. This set of masks is denoted as \( \mathcal{M}_{obj} = \{M_1, M_2, \ldots, M_n\} \), with each \( M_i \) representing a distinct object mask. To focus on the objects neighboring the query object mask \( M_{modal} \), we filter \( \mathcal{M}_{obj} \) to yield \( \mathcal{M}_{neighbor} = \{M_1, M_2, \ldots, M_j\} \).

We also perform a depth ordering analysis \cite{lee2022instance} between \( M_{modal} \) and the masks in \( \mathcal{M}_{neighbor} \), and we consider any mask in \( \mathcal{M}_{neighbor} \) that is closer to the camera than \( M_{modal} \) as an occluder. Next, this set of occluder masks \( \mathcal{M}_{occluder} \) is aggregated into a single binary occlusion mask, mathematically expressed as \( M_{occ} = \sum_{M_i \in \mathcal{M}_{occluder}} M_i \). This unified mask \( M_{occ} \) captures occluders within the image boundary and serves as the input mask for the diffusion process.

\textbf{Conditional padding.}
Our approach completes objects that may extend beyond the image boundary by including the boundary as an occluder.  If the query object mask \( M_{modal} \) touches the boundary, then we apply padding to the image \( I_{in} \) and the input mask \( M_{occ} \) in the corresponding directions.

\textbf{Diffusion process and occlusion check.}
After mask analysis and conditional padding, we zoom into the query object by cropping \( I_{in} \), \( M_{occ} \), and \( M_{modal} \) around its bounding box. This can improve the image generation quality by the diffusion inpainting process, as described in \cite{zhang2023perceptual_pal4vst}.

The input bundle to the diffusion process contains the new image \( I_{in} \), occluder mask \( M_{occ} \), query object's modal mask \( M_{modal} \), and semantic category for $P$. We run our Mixed Context Diffusion method (Section \ref{sec:method_mixed_context_sampling}) and generate a new amodal completion image \(I_{amodal}\) using the equation:

\vspace{-5 pt}
\begin{equation}
I_{amodal} = F_{0 \rightarrow N}(I_{in}, M_{occ}, P)
\label{equ:progressive_occlusion_aware_completion}
\end{equation}

At the end of each pipeline iteration, we check if the object is still occluded by other objects or the image boundary.

\textbf{Additional iterations.}
If occlusions remain, then we run another iteration of our pipeline using the previous iteration's amodal completion image \(I_{amodal}\) as the new input \( I_{in} \), and the previous iteration's amodal mask \( M_{amodal} \) as the new modal mask  \( M_{modal} \). Our pipeline continues until the query object is no longer occluded. Lastly, we return an output bundle with the final amodal completion image and amodal mask. To visualize the completed object in the original context, we can trim the extra background from \(I_{amodal}\) and overlay the object on the original image.

\subsection{Mixed Context Diffusion Sampling}
\label{sec:method_mixed_context_sampling}

Our Progressive Occlusion-aware Completion pipeline involves inpainting occluder regions, but directly using a pre-trained diffusion inpainting model may generate co-occurring objects as new occluders due to contextual bias. This bias extends to subtle details like shadows, which can prompt the model to produce contextually compatible occluders in the edited region, as discussed in \cite{zhan2023does}.

To address this, we temporarily break the co-occurrence link between the query object and original image context during the diffusion process. We achieve this through our `Mixed Context Diffusion Sampling' (MC), presented in Figure \ref{fig:mixed_context_sampling}. This approach is versatile and can be adapted to any text-to-image diffusion model in pixel or latent space.

After receiving the input bundle described in Section \ref{sec:method_progressive_completion_pipeline}, our approach bifurcates into two parallel paths. The first path aims to complete the query object by reducing contextual bias, while the second path frees the original image of occluders. Then, we composite the noisy images from both paths into a single noisy image by creating an intermediate query object mask. We explain each step below.

\textbf{Swap background.} We replace the area of \( I_{in} \) outside the query object’s modal mask \(M_{modal}\) with a clean background, reminiscent of the gray backdrops typically used in product photography. This creates a synthetically composited image, denoted as \( I_{syn} \). Next, in the Denoise step, we apply diffusion inpainting using \( I_{syn} \) and \( M_{occ} \) up to the \( k^{th} \) diffusion timestep. This can be mathematically expressed as:
\begin{equation}
I_{syn\_amodal}^k = F_{0 \rightarrow k}(I_{syn}, M_{occ}, P)
\label{equ:mc_synthetic_path}
\end{equation}

\textbf{Create object-removed background image.} We produce a clean background image of the original context, devoid of both query and occluder objects. To achieve this, we first remove them from \( I_{in} \) using the combined area of \( M_{modal} + M_{occ} \) via a removal inpainter \( R \), as described in \cite{suvorov2021lama}. The output from \( R \) is then subjected to noise addition using AddNoise$(\cdot, k)$ up to the \( k^{th} \) timestep. This results in \( I_{bg}^k \), which is the noise-infused clean background image after $k^{th}$ timesteps. This can be mathematically expressed as:
\begin{equation}
I_{bg}^k = \text{AddNoise}(R(I_{in}, M_{modal} + M_{occ}), k)
\label{equ:mc_remove_path}
\end{equation}

\textbf{Segment query object in noisy image.}
After deriving \( I_{syn\_amodal}^k \) and \( I_{bg}^k \), each characterized by the \( k^{th} \) noise level, we aim to insert the query object from \( I_{syn\_amodal}^k \) into the object-removed background image. Central to this step is the determination of an appropriate intermediate query object mask from the noisy image $I_{syn\_amodal}^k$.

Our insight is that segmenting the query object from the noisy image \( I_{syn\_amodal}^k \) is difficult, but we can use the latent information from the UNet decoder to find clusters \cite{kmeans_lloyd_algorithm} for query object mask proposals. We experimentally determined the best \( l^{th} \) decoder layer and \( k^{th} \) timestep to extract features. Each cluster is associated with different segments of the image. We compute pixel overlap of each cluster with the modal mask \( M_{modal} \) to select the segment that best aligns with the query object in the noisy image \( I_{syn\_amodal}^k \). This segment is the amodal object mask \( M_{amodal}^k \) at the \( k^{th} \) timestep.

\textbf{Composite.}
We use  \( M_{amodal}^k \) to composite the query object back onto the object-removed background image \( I_{bg}^k \), instead of the original image, to ensure that the completed query object is contextually consistent. We create this composited image \( I_{amodal}^k \) as follows:

\begin{equation*}
\begin{aligned}
I_{amodal}^k = \ &I_{syn\_amodal}^k \odot M_{amodal}^k \\
&+ I_{bg}^k \odot (1 - M_{amodal}^k)
\end{aligned}
\label{equ:mc_composite}
\end{equation*}

Finally, we continue the diffusion process for  \( N - k \) steps using the composited image  \( I_{amodal}^k \) and the occluder mask \( M_{occ} \). We denote this remaining diffusion process as \( F_{k \rightarrow N} \) and obtain the final image  \( I_{amodal} \). This is expressed as:

\vspace{-5 pt}
\begin{equation}
I_{amodal} = F_{k \rightarrow N}(I_{amodal}^k, M_{occ}, P)
\label{equ:mc_final_step}
\end{equation}
\vspace{-10 pt}

\subsection{Counterfactual Completion Curation System}

After generating a set of amodal completion images, how can we decide if the objects are successfully completed?
Inspired by \cite{Oktay2018CounterfactualIN}, we propose a `Counterfactual Completion Curation System' that can reduce the burden of human labeling by filtering unsuccessful completions without model training.  The intuition behind our system is that outpainting incomplete objects is more likely to generate more pixels belonging to missing object parts than outpainting complete objects. Our initial curation system relies on a training-free rule to classify generated objects as complete or incomplete.

\begin{figure}[H]
    \vspace{-6 pt}
    \centering
    \includegraphics[trim=0in 3.8in 2.95in 0in, clip,width=\columnwidth]{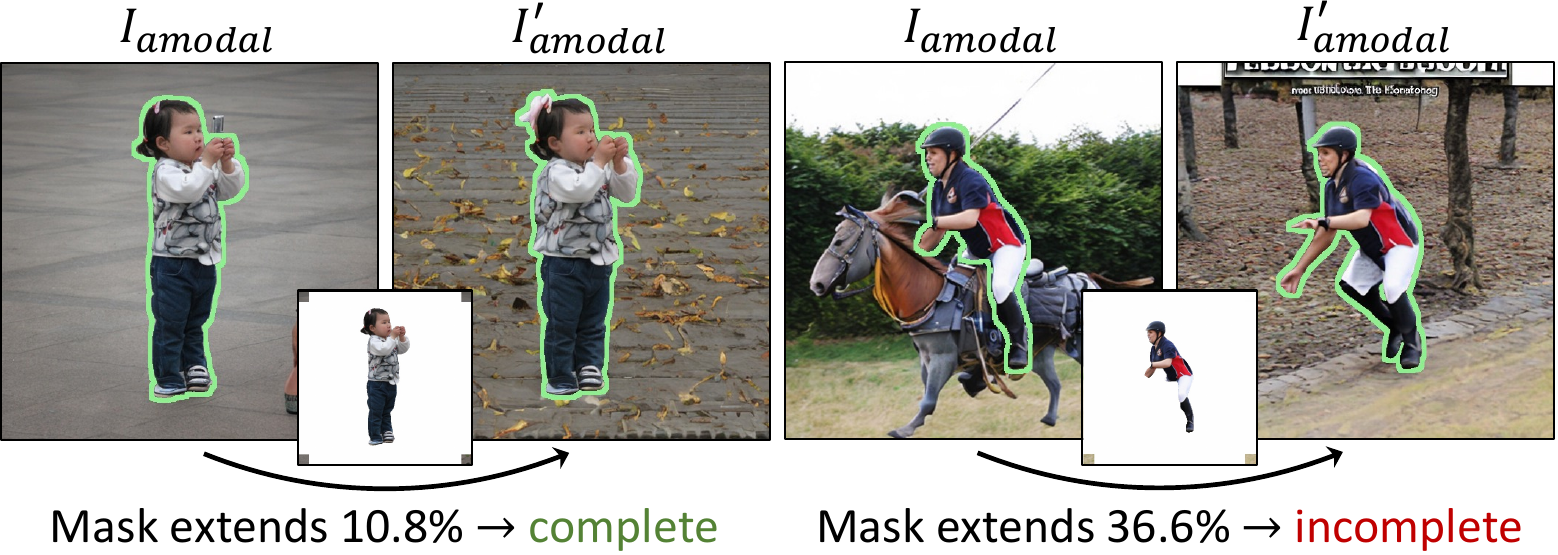}
    \vspace{-18 pt}
    \caption{Our counterfactual completion curation system uses a training-free rule to determine complete and incomplete objects. We outpaint the object everywhere except the image corners, and then compare amodal masks from \( I_{amodal} \) and \( I'_{amodal} \). We experimentally determined a mask extension threshold of 20\%.}
    \label{fig:counterfactual_rule}
    \vspace{-9 pt}
\end{figure}

Figure \ref{fig:counterfactual_rule} shows complete and incomplete objects determined by our rule, which has a generation step and a decision step. In the generation step, we outpaint the object in  \( I_{amodal} \) using an input mask \( M_{in} \) consisting of everywhere except the amodal mask \( M_{amodal} \) and the image corners, which guides the diffusion process towards a reasonable background.  Then, we trim the background in the new amodal completion image \( I'_{amodal} \) to extract the query object and obtain a new amodal mask \( M'_{amodal} \). In the decision step, we classify objects using thresholds on two parameters: 1) the object's proximity to the image boundary, and 2) the extension of the amodal mask area. To set the two thresholds, we use a small validation set of 100 images.

\section{Experiments}
\label{sec:experiments}

\begin{figure*}[!ht]
    \vspace{-15 pt}
    \centering
    \includegraphics[trim=0in 3.15in 1.1in 0in, clip,width=\textwidth]{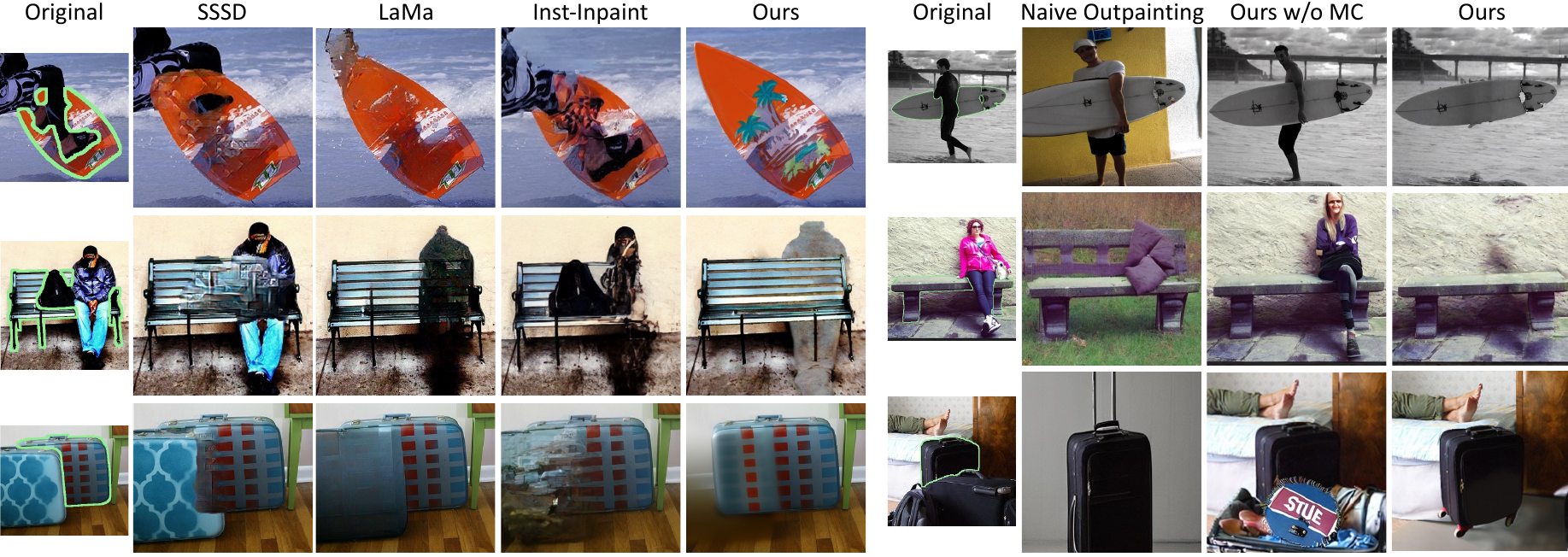}
    \vspace{-20 pt}
    \caption{\textbf{Left:} Comparison of our method with prior works on natural images. \textbf{Right:} Comparison of our method, our method without Mixed Context Diffusion Sampling (MC), and Naive Outpainting. Our method extends objects only where necessary unlike Naive Outpainting. Additionally, our approach avoids generating co-occurring objects, unlike ours without MC and Naive Outpainting.}
    \label{fig:experiments}
    \vspace{-12 pt}
\end{figure*}

Our Progressive Occlusion-aware Completion pipeline and Mixed Context Diffusion Sampling can successfully fill in hidden object pixels in a variety of object categories and occlusion cases in natural images. Notably, it can complete objects inside and outside the image boundary, overcome difficult co-occurrence bias, and handle high occlusion rates using the pre-trained diffusion model's good image prior.

\subsection{Comparisons with Previous Methods}

\textbf{Prior Works.} We compare with three prior works:
a GAN-based amodal completion method \emph{Self-Supervised Scene De-occlusion} (SSSD) \cite{zhan2020selfsupervised},
a GAN-based inpainting method \emph{Large Mask Inpainting} (LaMa) \cite{suvorov2021lama}, and
a diffusion-based object removal method \emph{Inst-Inpaint} \cite{yildirim2023instinpaint}. For fair comparison, we focus on completing objects within the image boundary because they do not extend the boundary. We use LaMa to fill in occluders and Inst-Inpaint to remove occluders.

\begin{figure}[!ht]
    \vspace{0 pt}
    \centering
    \includegraphics[trim=0in 4.9in 5.1in 0in, clip,width=\columnwidth]{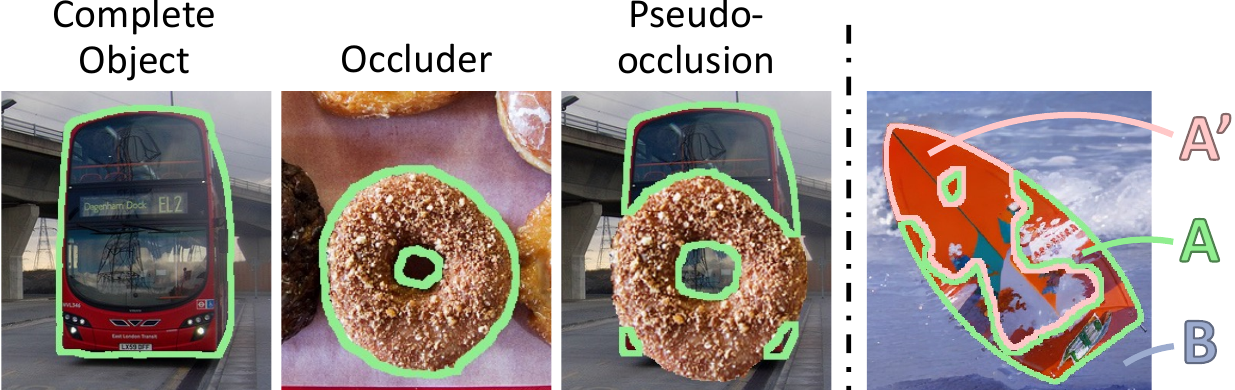}
    \vspace{-18 pt}
    \caption{\textbf{Left:} We place complete objects on each other to create pseudo-occluded objects. Here, the bus is 66.2\% occluded by the donut.  \textbf{Right:} We can evaluate whether $A'$ (formerly occluded) successfully completes the query object $A$ by using pseudo-occluded objects and off-the-shelf metrics. But, these metrics do not assess whether $A' \cup A$ fits into the background $B$. To this end, we conduct a user preference study to evaluate the generated objects in context.}
    \label{fig:pseudo_occlusion_creation}
    \vspace{-18 pt}
\end{figure}

\textbf{Datasets.} One main challenge of evaluating our amodal completion task is that there are no natural image datasets with ground truth amodal appearance completions for common object categories. In addition, existing amodal datasets with ground truth amodal masks do not consider the diversity of possible object completions. To bypass these limitations, we create a dataset of 3,000 pseudo-occluded common objects and their completed counterparts using natural images from COCO \cite{lin2015microsoft} and Open Images \cite{OpenImages, OpenImages2}. As shown in Figure \ref{fig:pseudo_occlusion_creation}, we simulate occlusion by overlaying a complete object on the image of another complete object. Our dataset contains diverse, challenging scenarios for amodal completion, covering at least 55 object categories with significant occlusion rates: 1,500 easy cases with 20-50\% occlusion and 1,500 hard cases with 50-80\% occlusion. In Figure \ref{fig:experiments}, we present qualitative results on natural images.

\textbf{Metrics.} We assess the quality of the generated amodal completion images with the ground truth complete object images at three image similarity levels. We use CLIP \cite{radford2021learning} for high-level, DreamSim \cite{fu2023learning} for mid-level, and LPIPS \cite{zhang2018unreasonable_lpips} for low-level. For CLIP, we compute the cosine similarity between image embeddings of the generated amodal completion and text embeddings of the query object category. For DreamSim and LPIPS, we calculate the perceptual distance between generated and ground truth complete object image. We segment and place the completed objects on a black background to focus on the query object appearance. Furthermore, we conduct a user preference study to evaluate how well the generated object fits into its original context. The right side of Figure \ref{fig:pseudo_occlusion_creation} provides an example of the various aspects of the amodal completion image to evaluate.

\textbf{Quantitative Results.} In Table \ref{tab:experiment_comparisons}, we report the mean image similarity scores across all pseudo-occluded object images. Our method generally performs better than prior works for both easy cases and hard cases. Interestingly, LaMa \cite{suvorov2021lama} obtains similar scores to our method even though it is not intended for amodal completion. We suspect that LaMa seems to perform well because it often generates similarly colored pixels as the query object within the inpainted region, and a visual assessment of LaMa's output verifies that it creates blurry object appearances and boundaries. For this reason, we conduct a user preference study and present qualitative results to further measure the amodal completion quality between our method and prior works.

\begin{table*}[!ht]
\vspace{-15 pt}
\centering
\resizebox{\textwidth}{!}{%
\begin{tabular}{@{\extracolsep{6pt}}lccccccc@{}}
\toprule
\multirow{2}{*}{Method} & \multicolumn{3}{c}{Easy Cases} & \multicolumn{3}{c}{Hard Cases} &        \multirow{2}{*}{User Preference} \\ \cline{2-4} \cline{5-7}
             &  CLIP $\uparrow$                & DreamSim $\downarrow$           & LPIPS $\downarrow$              & CLIP $\uparrow$        & DreamSim $\downarrow$           & LPIPS $\downarrow$              & \\ \midrule
SSSD \cite{zhan2020selfsupervised}         & 0.280 / 0.263                   & 0.186 / 0.216                   & 0.096 / 0.142                   & 0.267 / 0.263          & 0.315 / 0.334                   & 0.166 / 0.225                   & 1.8\%           \\ 
LaMa \cite{suvorov2021lama}         & 0.288 / 0.265                   & 0.098 / 0.124                   & \textbf{0.054} / 0.091          & 0.279 / \textbf{0.268} & 0.236 / 0.292                   & 0.130 / 0.205                   & 7.3\%           \\ 
Inst-Inpaint \cite{yildirim2023instinpaint} & 0.264 / 0.257                   & 0.325 / 0.304                   & 0.185 / 0.195                   & 0.252 / 0.254          & 0.451 / 0.446                   & 0.263 / 0.283                   & 0.0\%             \\
Ours         & \textbf{0.290} / \textbf{0.266} & \textbf{0.096} / \textbf{0.106} & \textbf{0.054} / \textbf{0.078} & \textbf{0.290} / 0.267 & \textbf{0.184} / \textbf{0.185} & \textbf{0.110} / \textbf{0.141} & \textbf{90.9\%} \\ \toprule
\end{tabular}
}%
\vspace{-8 pt}
\caption{Our method overall performs better than prior works in terms of CLIP (high-level), DreamSim (mid-level), and LPIPS (low-level) image similarity. We use 2,500 objects from COCO and 500 objects from Open Images, and \emph{scores are formatted as COCO / Open Images.} We consider easy cases where the object has 20-50\% occlusion and hard cases with 50-80\% occlusion. Additionally, we observe that users highly prefer the generated amodal completions using our method across 55 easy cases and 55 hard cases from COCO.}
\label{tab:experiment_comparisons}
\vspace{-12 pt}
\end{table*}

\textbf{User Preference Study.} We conduct a user preference study to more accurately assess the perceptual image quality of the generated amodal completions. We randomly select 55 easy cases and 55 hard cases of pseudo-occlusion images and solicit feedback from at least three Amazon Mechanical Turk (MTurk) workers. We show the pseudo-occluded object image and the generated object images from each method side by side, and we ask each worker to vote on the generated object that looks most complete and realistic. As shown in Table \ref{tab:experiment_comparisons}, the user preferences demonstrate that our method significantly outperforms prior works in the visual quality of amodal completion images.

\textbf{Qualitative Results.} Figure \ref{fig:experiments} visually compares our method and prior works on occluded objects in natural images. We notice that SSSD \cite{zhan2020selfsupervised} often generates visual artifacts and incomplete objects, LaMa \cite{suvorov2021lama} produces ill-defined object boundaries and unrealistic appearances, and Inst-Inpaint \cite{yildirim2023instinpaint} can remove objects but struggles to complete them. Our method can complete highly occluded objects with realistic and contextually consistent appearances.

\textbf{Implementation.} We demonstrate our method using the publicly available Stable Diffusion v2 inpainting model \cite{rombach2022ldm}. All experiments use a 24GB Nvidia Titan RTX GPU, and our method does not involve any training or fine-tuning.

\subsection{Ablation Studies}
\label{sec:experiments_ablation_studies}

We ablate our amodal completion method to demonstrate the effectiveness of our Progressive Occlusion-aware Completion pipeline and Mixed Context Diffusion Sampling. We randomly select 100 occluded objects from natural images and generate their completed versions using our method and Naive Outpainting. We consider 50 hard cases where the occluder is the top co-occurring semantic category for the query object, and 50 easy cases where the occluder is not.

We conduct a user study to find the number of successful amodal completions for each method by soliciting feedback from at least three MTurk workers. In addition, we perform a user preference study and ask at least six MTurk workers to vote on the method that generates the most complete and realistic objects. In Table \ref{tab:ablation_user_study}, our method outperforms Naive Outpainting in successful completions and user preference, even without Mixed Context Diffusion Sampling. On hard cases, we observe that using Mixed Context Diffusion Sampling significantly aids successful completions by +18\%.

\begin{table}[!h]
\centering
\vspace{0 pt}
\resizebox{\columnwidth}{!}{%
\begin{tabular}{@{\extracolsep{4pt}}lcccc@{}}
\toprule
\multirow{2}{*}{Method} & \multicolumn{2}{c}{Easy Cases} & \multicolumn{2}{c}{Hard Cases} \\ \cline{2-3} \cline{4-5}
             & Successes     & User Preference & Successes     & User Preference \\ \midrule
Naive Outpainting     & 66\%          & 18\%            & 40\%          & 18\%            \\ 
Ours w/o MC  & \textbf{90\%} & 36\%            & 72\%          & 28\%            \\ 
Ours         & 88\%          & \textbf{48\%}   & \textbf{90\%} & \textbf{54\%}   \\ \toprule
\end{tabular}
}%
\vspace{-7 pt}
\caption{Ablation study of amodal completion successes and user preference. We use 50 easy cases and 50 hard cases, where the occluder is the top co-occurring object category.}
\label{tab:ablation_user_study}
\vspace{-12 pt}
\end{table}

Figure \ref{fig:experiments} visually compares each method on natural images with challenging co-occurrence or highly occluded objects. Our method prevents co-occurring objects and inpaints only where necessary, compared to ours without Mixed Context Diffusion Sampling and Naive Outpainting.

\subsubsection{Mixed Context Diffusion Sampling}

We experiment with the clean image to swap background, the UNet layer to cluster features and segment the query object in the noisy image, and the timestep to composite the query object on the object-removed background image.

\begin{figure}[!ht]
    \vspace{-5 pt}
    \centering
    \includegraphics[trim=0.03in 5.6in 7.25in 0in, clip,width=\columnwidth]{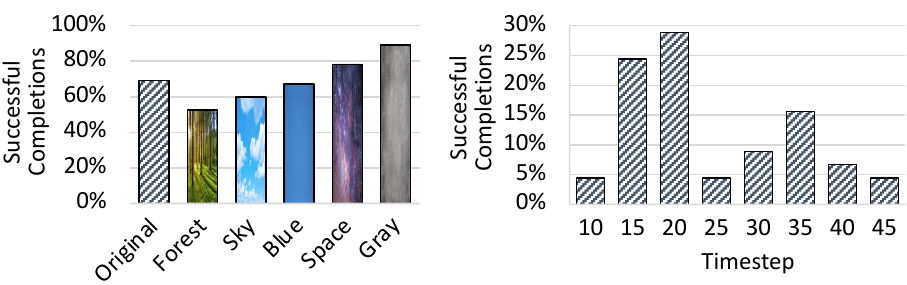}
    \vspace{-18 pt}
    \caption{\textbf{Left:} Using a gray background improves completion by +20\% compared to the original background.  \textbf{Right:} Swapping contexts at DDIM timestep 20 out of 50 leads to more successful completions on \emph{difficult co-occurrence cases}.}
    \label{fig:mixed_context_ablation_graphs}
    \vspace{-8 pt}
\end{figure}

\textbf{Clean image to swap background.}
We test the effect of five clean backgrounds to swap with the original image background on a small set of 55 images. On the left of Figure \ref{fig:mixed_context_ablation_graphs}, using a gray background led to a +20\% increase in successful amodal completions, while using a forest or sky background led to a noticeable drop in completion performance (-16\% and -9\%) compared to using the original context.

\begin{figure}[!t]
    \vspace{-7 pt}
    \centering
    \includegraphics[trim=0in 7.4in 4.95in 0in, clip,width=\columnwidth]{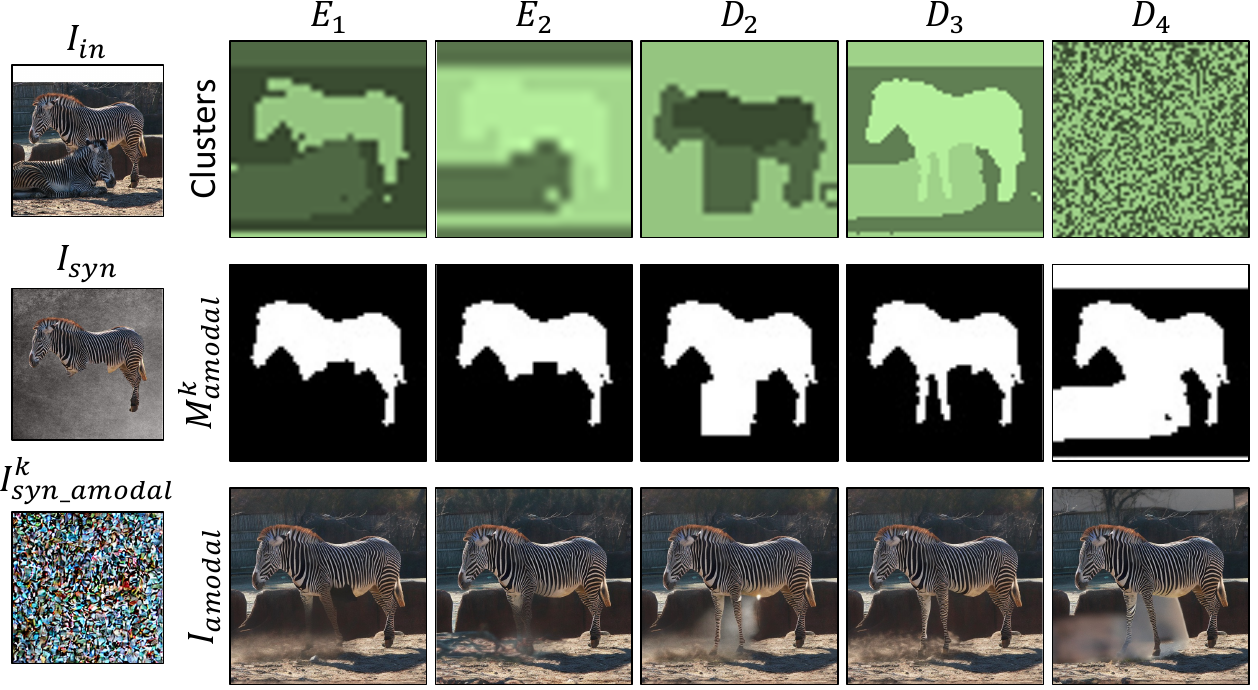}
    \vspace{-18 pt}
    \caption{We analyze how features from different UNet encoder and decoder layers affect the clusters, intermediate amodal mask \( M_{amodal}^k \), and final amodal completion image \( I_{amodal} \). Here, we show a subset of the UNet layers, and $E_l$ and $D_l$ refer to the $l$-th layer of the encoder and decoder, where $l \in [1, 2, 3, 4]$. The leftmost column shows the input image \( I_{in} \) for the current iteration of our pipeline, the synthetically overlaid input image \( I_{syn} \), and the intermediate inpainted object \( I_{syn\_amodal}^k \). We discover that features from $D_3$ generally produce good clusters to create a well-defined \( M_{amodal}^k \). Features from other layers often cannot fully capture thin structures, such as the zebra's legs in this example.}
    \label{fig:mixed_context_sampling_ablation}
    \vspace{-18 pt}
\end{figure}

\textbf{UNet layer to segment object in noisy image.}
To cluster features and create the intermediate amodal mask \( M_{amodal}^k \), we examine the effect of using different UNet layers in Figure \ref{fig:mixed_context_sampling_ablation}. We observe that features from the third UNet decoder layer best capture object geometry and low-level visual features for Mixed Context Diffusion Sampling, as described in \cite{tang2023emergent, mou2023dragondiffusion}. Early encoder layers sometimes cluster the inpainting mask region, and the last decoder layer is often noisy. The second decoder layer sometimes captures the general object shape but is less fine-grained. Despite poor clusters and intermediate amodal mask, the final amodal completion of the query object is not always negatively impacted.

\textbf{Timestep to composite.}
On the right of Figure \ref{fig:mixed_context_ablation_graphs}, we analyze the timestep to swap image backgrounds on 45 images where our method without Mixed Context Diffusion Sampling generates a co-occurring occluder or unwanted visual artifacts. We record the earliest timestep that returns a good intermediate amodal mask \( M_{amodal}^k \) and final amodal completion image \( I_{amodal} \). We observe the peak at timestep 20, indicating that object shape may appear relatively early on during the denoising process. Nonetheless, the ideal timestep for compositing the query object onto the object-removed background image often depends on the occlusion level.

\subsection{Counterfactual Completion Curation System}

Our curation system involves a training-free rule that classifies generated objects as complete or incomplete.  On the validation set of 100 images, our rule reaches a mean accuracy of 0.73, precision of 0.73, and recall of 0.71 from 3 different trials. We evaluate our rule by measuring accuracy, precision, and recall on a test set of 50 complete objects and 50 incomplete objects. As shown in Table \ref{tab:counterfactual_exp}, our rule can achieve an accuracy of 0.70, despite not being specially trained on any datasets. There is room for improvement, which can be mitigated by training models on a curated dataset of complete and incomplete objects.

Furthermore, we compare the performance of our rule with that of humans. We ask three humans to independently classify the 100 test images, and we compute the human consensus using a simple majority vote for each object. Human consensus results in an accuracy of 83\%. This indicates the subjective nature of this binary classification task and further shows the need for an accurate and reliable curation system.

\begin{table}[!ht]
\centering
\vspace{-7 pt}
\resizebox{0.85\columnwidth}{!}{%
\begin{tabular}{lccc}
\toprule
                 & Accuracy & Precision & Recall \\ \hline
Human Consensus  & 0.83     & 0.75      & 0.99  \\ 
Our Rule         & 0.70     & 0.68      & 0.68  \\
Random Chance    & 0.50     & 0.50      & 0.50  \\ \toprule
\end{tabular}
}%
\vspace{-8 pt}
\caption{Comparison of our counterfactual rule with humans on classifying 100 test images as complete or incomplete. We compare the mean scores of our rule from three different trials with the human consensus as a simple majority vote from three humans.}
\label{tab:counterfactual_exp}
\vspace{-15 pt}
\end{table}

\section{Discussion}
\label{sec:discussion}

\begin{figure}[!h]
    \vspace{-7 pt}
    \centering
    \includegraphics[trim=0in 5.1in 8.7in 0in, clip,width=\columnwidth]{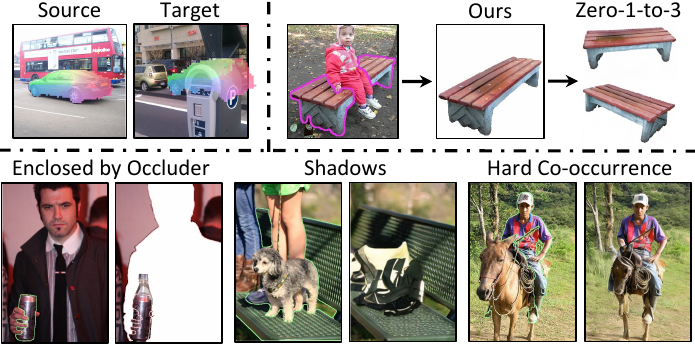}
    \vspace{-24 pt}
    \caption{\textbf{Top:} Two applications of our amodal completion method are dense correspondence \cite{zhang2023tale_sd_dino} and novel view synthesis \cite{liu2023zero1to3}.  \textbf{Bottom:} One failure case occurs when the object is enclosed by the occluder, leading to overextension. Additionally, our method sometimes struggles to complete objects due to shadows or the object's pose, such as the person riding a horse.}
    \label{fig:applications}
    \vspace{-8 pt}
\end{figure}

We introduced a new approach for amodal completion using diffusion inpainting. Our method progressively inpaints obscured regions by analyzing occlusion for a query object. Unlike conventional two-step approaches that predict the amodal mask and then complete the amodal appearance, ours is the first to predict the appearance directly. We deploy mixed context diffusion sampling to reduce the inpainting of unintended co-occurring objects. We also establish a counterfactual-based curation system for measuring object completeness. With our method, we can build dense correspondence \cite{zhang2023tale_sd_dino} and 3D novel view synthesis \cite{liu2023zero1to3} on highly occluded visual objects, as shown in Figure \ref{fig:applications}.

\textbf{Limitations.} Our method, while effective, encounters limitations with everyday occlusions, as illustrated in Figure \ref{fig:applications}. Challenges arise when a small query object is obscured by a larger occluder, potentially leading to its overextension. Additionally, subtle shadows on the query object can inadvertently introduce compatible occluders. Moreover, certain human poses strongly suggest interaction with other objects, occasionally resulting in the generation of unintended occluders. Despite these challenges, we believe our method establishes a solid new benchmark in amodal completion. Addressing these complex scenarios remains an exciting avenue for future work.

{\small
\bibliographystyle{ieeenat_fullname}
\bibliography{11_references}
}

\ifarxiv \clearpage \appendix \section{Implementation Details}

\subsection{Progressive Occlusion-aware Completion}

To recover unoccluded appearances of objects, our method inpaints necessary regions by identifying occluders and iteratively performs this inpainting step to avoid incompletion.

\textbf{Mask analysis.}
The first step of each iteration is performing instance segmentation \cite{liu2023groundingdino, kirillov2023segment_sam} and analyzing the object masks to determine occluders. Given the modal mask \( M_{modal} \) of a query object, we find its neighboring masks \( \mathcal{M}_{neighbor} \). Then, we perform a depth ordering analysis and filter \( \mathcal{M}_{neighbor} \) to contain masks closer to the camera than \( M_{modal} \). This gives a refined set of masks, \( \mathcal{M}_{occluder} \), which is aggregated into a single binary occlusion mask \( M_{occ} \). Figure \ref{fig:supp_mask_analysis} shows additional examples of this mask analysis step.

\begin{figure}[H]
    \vspace{-6 pt}
    \centering
    \includegraphics[trim=0in 2.5in 8.05in 0in, clip,width=0.95\columnwidth]{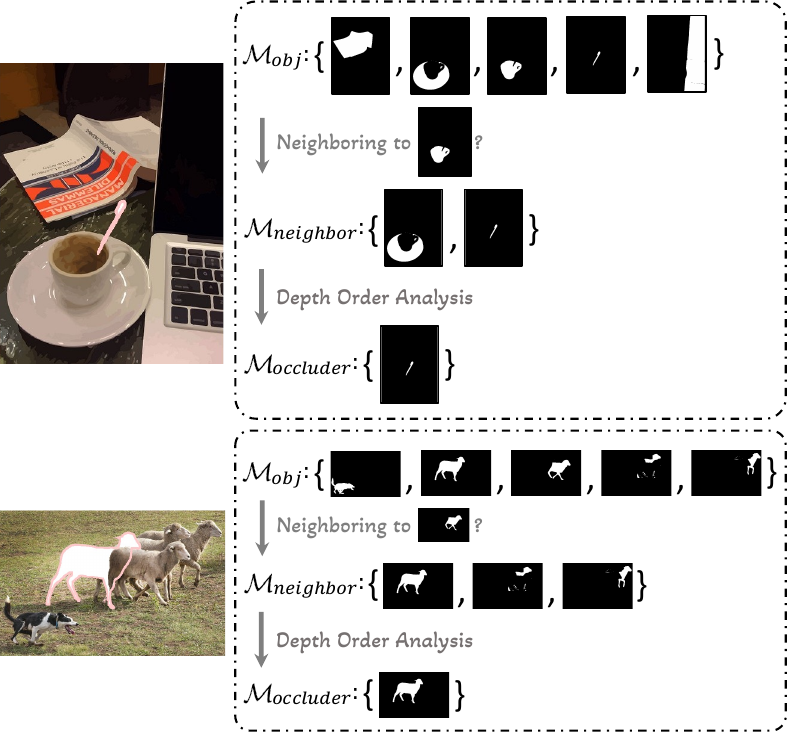}
    \vspace{-10 pt}
    \caption{Mask analysis examples. \textbf{Top:} The cup is occluded by a spoon. \textbf{Bottom:} The sheep is occluded by another sheep.}
    \label{fig:supp_mask_analysis}
    \vspace{-10 pt}
\end{figure}

\textbf{Conditional padding.}
If the query object lies within 10 pixels of the image boundary, then we pad the image \( I_{in} \) and mask \( M_{occ} \) with white pixels in those directions.

\textbf{Diffusion process and occlusion check.}
To create the input bundle, we crop the image \( I_{in} \) and mask \( M_{occ} \) into squares around the query object. First, we extend the object's tight bounding box by $\alpha$ pixels on each side. Second, if the object touches the image boundary, then we extend the bounding box by an additional $\beta$ pixels on each side. Third, if the bounding box is not a square, then we extend the two shorter sides to have the same length as the longer sides. We use $\alpha=60$, $\beta=60$ in our experiments.

Moreover, we identify new occlusions by repeating the segment and mask analysis steps, as well as checking if the query object lies within $\lambda$ pixels of the image boundary. We use $\lambda=10$ to account for potentially inaccurate segmentations from the grounded segmentation models \cite{liu2023groundingdino, kirillov2023segment_sam} on the image boundary.

\subsection{Mixed Context Diffusion Sampling}

We describe the details of segmenting the query object in the noisy image \( I^{k}_{syn\_amodal} \).

\textbf{Extract features.}
At the \( k^{th} \) DDIM timestep, we extract text-conditioned features from the \( l^{th} \) UNet decoder layer.  We experimentally determined that $l=3$ is favorable for capturing an object's shape and $k=20$ is generally better for achieving successful completions on difficult co-occurrence cases, but we acknowledge that the precise layer and timestep often depends on the occlusion level.

\textbf{Cluster features.}
The features from the third decoder layer are flattened and permuted from size \( (640, 64, 64) \) to \( (64 \times 64, 640) \). We perform unsupervised clustering \cite{kmeans_lloyd_algorithm} of the features and reshape the cluster assignments to \( (64, 64) \).

\textbf{Align clusters with query object.}
To segment the object in a noisy image, we upsample the cluster assignments to the size of the query object's modal mask, and then select the clusters with more than $20\%$ overlap with the modal mask. Additionally, we include the modal mask in the segmentation, and we constrain the segmentation to only regions that are within the modal mask or occluder mask.

\subsection{Counterfactual Completion Curation System}

Our training-free rule outpaints generated objects and classifies them as complete or incomplete using two parameters: 1) the object's proximity to the image boundary in \( I'_{amodal} \), and 2) the extension of the amodal mask area from \( M_{amodal} \) to \( M'_{amodal} \).
First, if any object pixel in \( M'_{amodal} \) is within $\gamma$ pixels of the image boundary, then we judge the object as incomplete due to potential occlusion.
Second, we dilate \( M_{amodal} \) using a $5 \times 5$ kernel and $\delta$ iterations. If \( M'_{amodal} \) is contained in the dilated \( M_{amodal} \), then we judge the object as complete to allow minor extensions of complete objects. If \( M'_{amodal} > \epsilon M_{amodal} \), then we judge the object as incomplete due to major extension.
We experimentally determined $\gamma=2, \delta=4, \epsilon=1.2$. Figure \ref{fig:supp_counterfactual} shows additional examples of complete and incomplete objects determined by our rule.

\section{Additional Qualitative Results}

We use our method to complete objects in natural images. Figure \ref{fig:supp_qual_results} illustrates amodal completions within and beyond the image boundary. Figure \ref{fig:diverse_completions} presents diverse versions of completed objects. Figure \ref{fig:supp_comparisons} compares our method with prior works. Figure \ref{fig:supp_ablations} shows comparisons with Naive Outpainting and ours without Mixed Context Diffusion.

\section{Experiment Details}

\subsection{User Studies}

\textbf{User preference study.}
For each study, we place the generated object from each method side by side, and then ask MTurk workers to select the object that looks most complete and realistic given the original image.
Humans must correctly answer five out of six attention check images.

For comparisons with prior works, we asked at least three humans to label each batch of images, with a total of 110 images in four batches.
For comparisons with Naive Outpainting and our method without Mixed Context Diffusion Sampling, we asked at least six humans to label each batch of images, with a total of 100 images in two batches.

\textbf{User study for successful completions.}
We randomly select 100 occluded objects in natural images and generate their amodal completions using each method. We ask at least three MTurk workers to label each batch with around 50 images. Humans are given three choices: complete object, incomplete/overextended object, or discard if they are unsure. They must correctly answer four out of five attention check images. Figure \ref{fig:mturk_ablation_success_example} shows instructions displayed to users.

\textbf{Human consensus for amodal completion.}
We asked three humans to judge the generated object in each amodal completion image as complete or incomplete. The human consensus is determined by a simple majority vote.

\begin{figure}[!t]
    \vspace{-15 pt}
    \centering
    \includegraphics[trim=0in 1in 8.85in 0in, clip,width=\columnwidth]{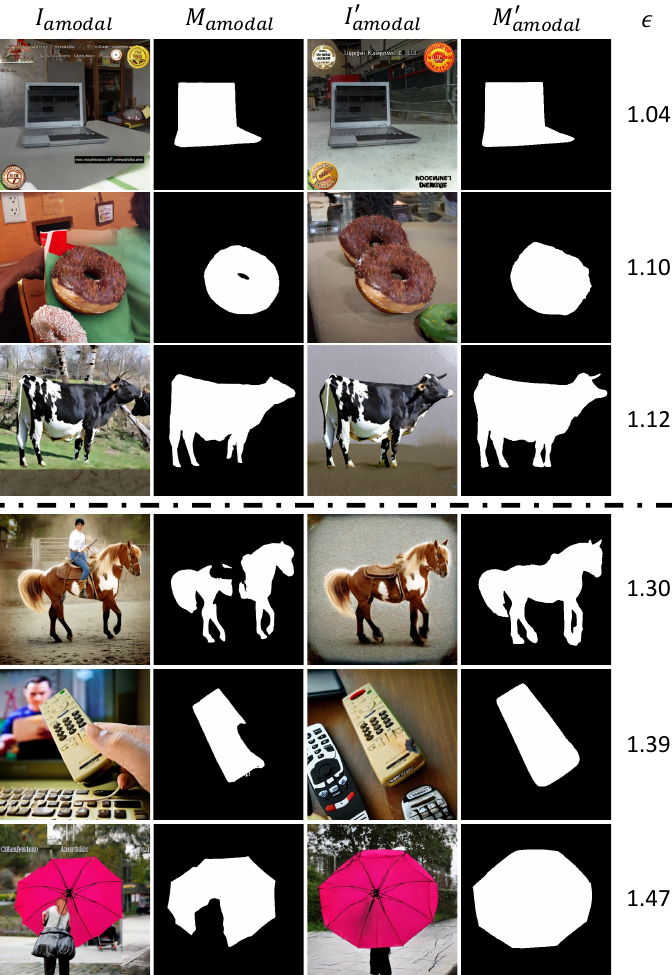}
    \vspace{-17 pt}
    \caption{Complete and incomplete objects predicted by our counterfactual rule. If the new object mask \( M'_{amodal} \) is greater than $\epsilon=1.2$ times the previous object mask \( M_{amodal} \), then the object is predicted as incomplete. \textbf{Top:} Complete. \textbf{Bottom:} Incomplete.}
    \label{fig:supp_counterfactual}
    \vspace{-10 pt}
\end{figure}

\begin{figure*}[!h]
    \vspace{-5 pt}
    \centering
    \includegraphics[trim=0in 4.6in 1.35in 0in, clip,width=0.7\textwidth]{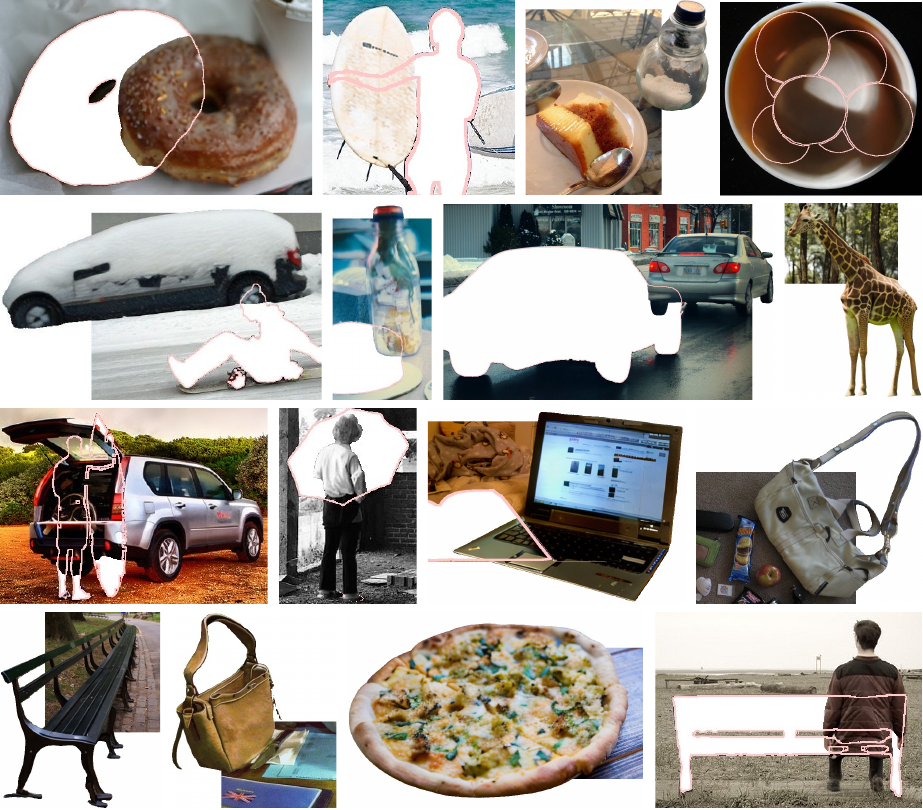}
    \vspace{-5 pt}
    \caption{Our method completes objects within and beyond the image boundary.}
    \label{fig:supp_qual_results}
    \vspace{-5 pt}
\end{figure*}

\begin{figure*}[!h]
    \vspace{-15 pt}
    \centering
    \includegraphics[trim=0in 4.98in 3.15in 0in, clip,width=\textwidth]{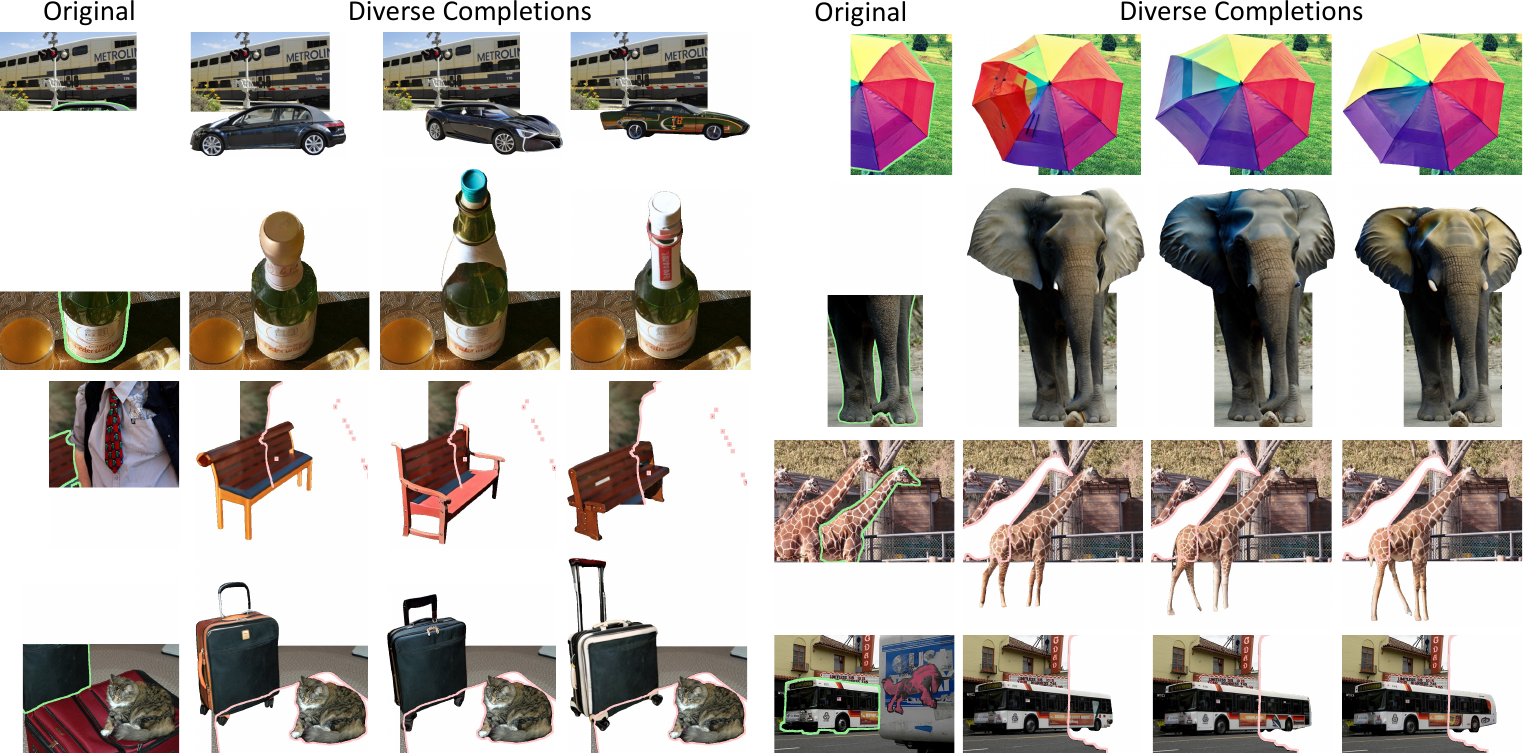}
    \vspace{-15 pt}
    \caption{We can obtain diverse amodal completions for each occluded object.}
    \label{fig:diverse_completions}
    \vspace{-5 pt}
\end{figure*}

\begin{figure*}[!h]
    \vspace{-10 pt}
    \centering
    \includegraphics[trim=0in 1in 5.1in 0in, clip,width=0.75\textwidth]{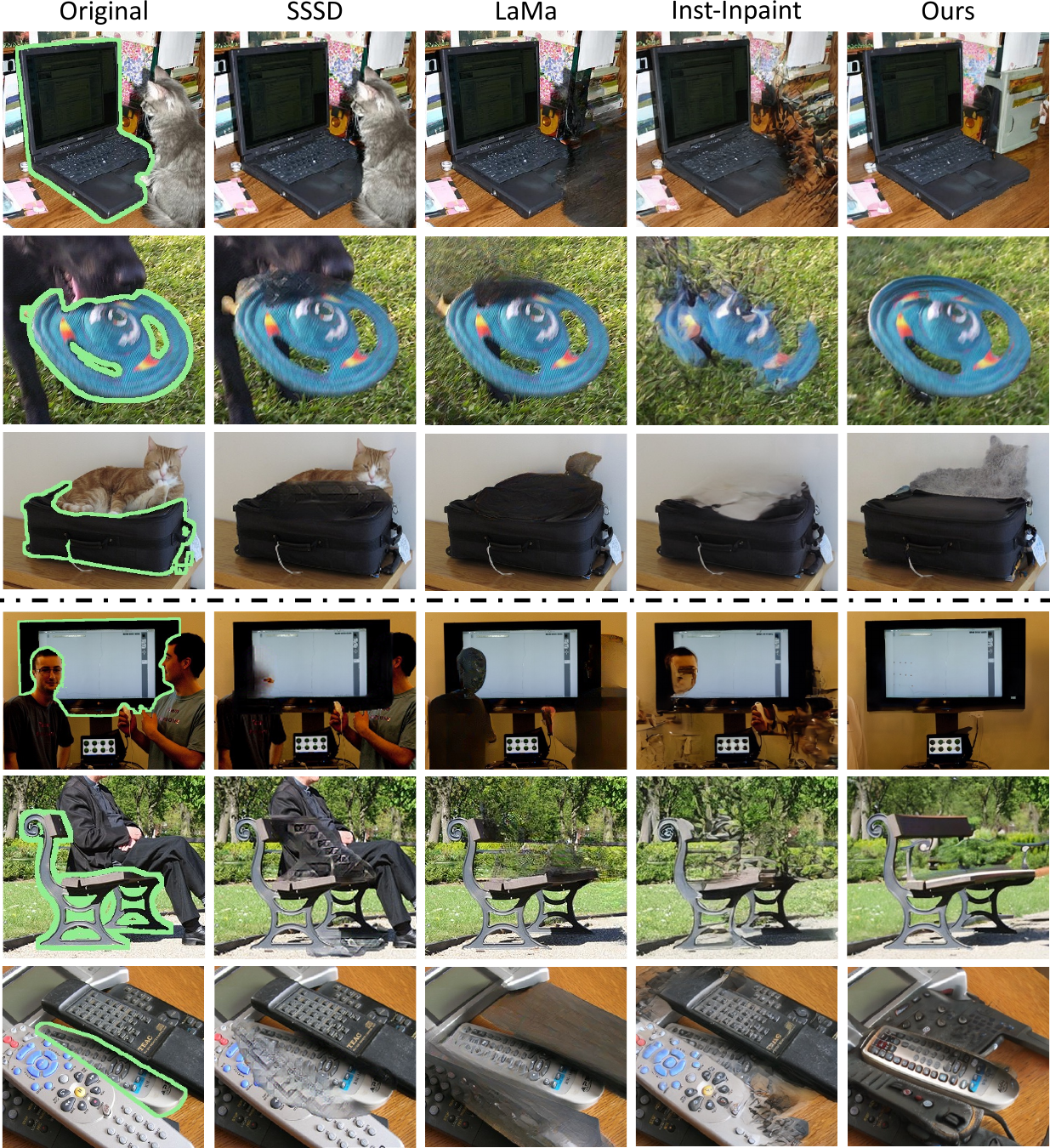}
    \vspace{-5 pt}
    \caption{Comparisons with prior works. \textbf{Top:} Easy cases. \textbf{Bottom:} Hard cases where the occluder is the top co-occurring object category for the query object. We find co-occurrence by randomly sampling 15,000 COCO images \cite{lin2015microsoft} and analyzing objects that appear together.}
    \label{fig:supp_comparisons}
    \vspace{-10 pt}
\end{figure*}

\begin{figure*}[!h]
    \vspace{-15 pt}
    \centering
    \includegraphics[trim=0in 4in 7.1in 0in, clip,width=0.7\textwidth]{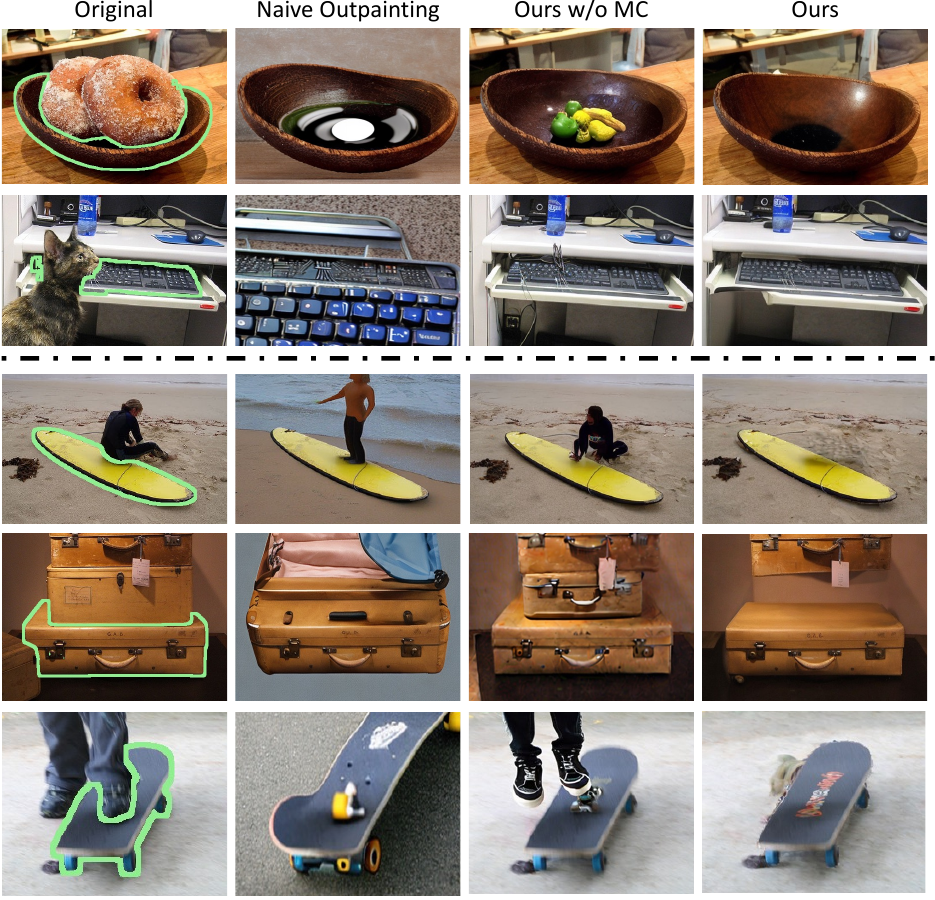}
    \vspace{-6 pt}
    \caption{Comparisons with Naive Outpainting and our method without Mixed Context (MC) Diffusion Sampling. \textbf{Top:} Easy cases. \textbf{Bottom:} Hard cases where the occluder is the top co-occurring object category for the query object.}
    \label{fig:supp_ablations}
    \vspace{0 pt}
\end{figure*}

\begin{figure*}[!h]
    \vspace{0 pt}
    \centering
    \includegraphics[trim=0in 5.5in 0in 0in, clip,width=0.9\textwidth]{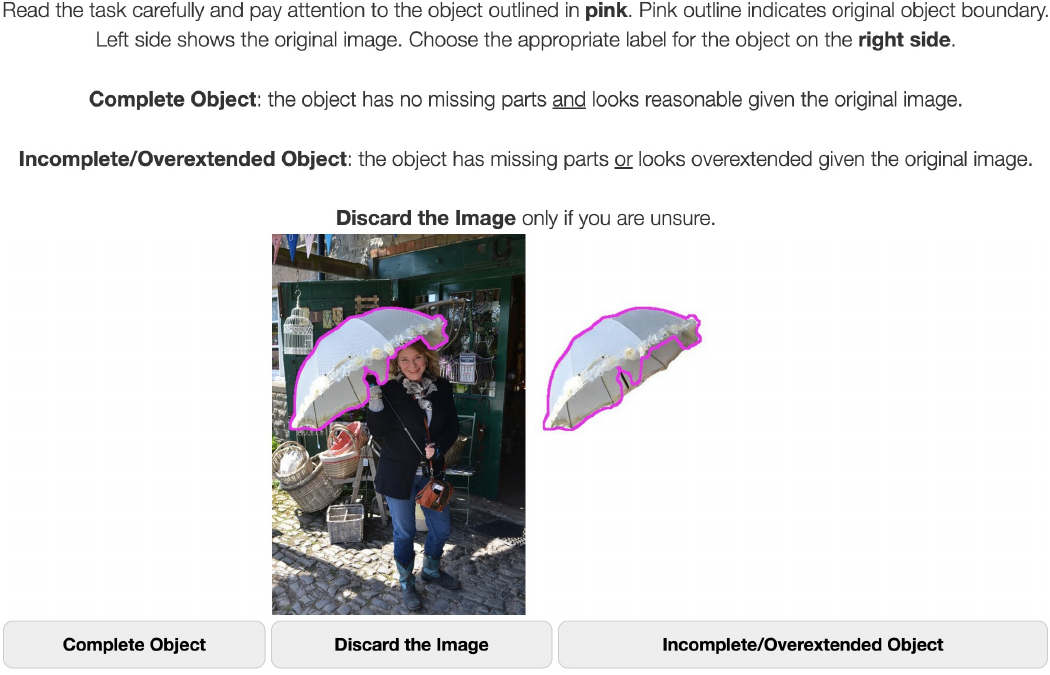}
    \vspace{-6 pt}
    \caption{Example set of instructions displayed to MTurk workers for our user study to determine successful completions.}
    \label{fig:mturk_ablation_success_example}
    \vspace{-5 pt}
\end{figure*}

 \fi

\end{document}